\documentclass[journal]{IEEEtran}
\usepackage{amsfonts}
\usepackage{amssymb}
\usepackage{algorithm2e}
\usepackage{algorithmic}
\usepackage{pgfplots}
\usepackage{mathtools}
\usepackage{algorithm2e}
\usepackage{algorithmic}
\usepackage{pgfplots}
\usepackage{amsfonts}
\usepackage{amssymb}
\usepackage{color}
\usepackage{algorithm2e}
\usepackage{algorithmic}
\usepackage{pgfplots}
\usepackage{amsfonts}
\usepackage{amssymb}
\usepackage{mathtools}
\usepackage{bbm}
\usepackage{dsfont}
\usepackage{float}

\usepackage{amsfonts}
\usepackage{amssymb}
\usepackage{algorithm2e}
\usepackage{algorithmic}
\usepackage{pgfplots}
\usepackage{mathtools}
\usepackage{color}
\usepackage{algorithm2e}
\usepackage{algorithmic}
\usepackage{pgfplots}
\usepackage{amsfonts}
\usepackage{amssymb}
\usepackage{mathtools}
\usepackage{bbm}
\usepackage{dsfont}
\usepackage{float}
\usepackage{hyperref}
\usepackage{comment}

\ifCLASSINFOpdf
\else
\fi
\begin{document}
%
\title{Gaussian Smoothen Semantic Features (GSSF) - Exploring the Linguistic Aspects of Visual Captioning in Indian Languages (Bengali) Using MSCOCO Framework} 
%
%
%

\author{Chiranjib~Sur \\ 
		Computer \& Information Science \& Engineering Department, University of Florida.\\
		Email: chiranjibsur@gmail.com
}

%
%

\markboth{Journal of XXXX,~Vol.~XX, No.~X, AXX~20XX}%
{Shell \MakeLowercase{\textit{et al.}}: Bare Demo of IEEEtran.cls for IEEE Journals}
%

\maketitle

\begin{abstract}
In this work, we have introduced Gaussian Smoothen Semantic Features (GSSF) for Better Semantic Selection for Indian regional language-based image captioning and introduced a procedure where we used the existing translation and English crowd-sourced sentences for training. We have shown that this architecture is a promising alternative source, where there is a crunch in resources. Our main contribution of this work is the development of deep learning architectures for the Bengali language (is the fifth widely spoken language in the world) with a completely different grammar and language attributes. We have shown that these are working well for complex applications like language generation from image contexts and can diversify the representation through introducing constraints, more extensive features, and unique feature spaces. We also established that we could achieve absolute precision and diversity when we use smoothened semantic tensor with the traditional LSTM and feature decomposition networks. With better learning architecture, we succeeded in establishing an automated algorithm and assessment procedure that can help in the evaluation of competent applications without the requirement for expertise and human intervention. 
\end{abstract}


%
\IEEEpeerreviewmaketitle

\section{Introduction} \label{section:introduction}
\IEEEPARstart{I}{mage} captioning \cite{sur2019survey} applications have gained massive attention from the research community both in academia and industry due to its ability to find a confluence of media and language and create the capability for machines to communicate in different languages with the external world. Image captioning has a wide range of applicability to tag enormous resources in the form of images \cite{Karpathy2015Deep} and video, and vast amounts of new media contents are added daily, which are un-categorized, and no one knows what its content is. This kind of image captioning brings these media in the mainstream web system through the use of these machines learned and automated infrastructures and applications.
While general object detection models introduced huge improvement like ResNet (\cite{Devlin2015Language, Gan2016}), GoogleNet \cite{vinyals2015show}, Vgg \cite{Karpathy2015Deep}, through the use of ImageNet \cite{fu2018image} data sources, there is need to establish the linkage and interaction among the different objects \cite{anderson2018bottom} and establish in sentences. While a large part of the resources is available in English languages, many regional languages are neglected, while large investments for these languages are not viable options. While image captioning is a complex task and gathering improvement is difficult, mainly because machines still do not get acquainted with the language attributes \cite{Gan2016} and grammar \cite{vinyals2015show}, we experiment with the viability of machine translation as an alternative. However, the use of machines is not to outsource the best practices of literary works, but to convey and interpret the messages without ambiguity. We discuss different existing image captioning architectures and translation procedures in the subsequent sections. 
Previous works in image captioning were relied on object and attribute detectors to describe images (\cite{Karpathy2015Deep, Chen2015Mind}) and later mostly concentrated on attention (\cite{Xu2015Show, vinyals2015show, Mao2014deep, Devlin2015Language, yao2017boosting, rennie2017self, chen2018show}) and compositional characteristics \cite{Gan2016} and top-down compositions \cite{anderson2018bottom}. At the same time, there were limited works on feature refinement and redefining the exiting subspace with more effective and efficient ones. \cite{lu2018neural} claimed to provide template-based sentence fill up, but no details are provided for the criteria of how objects are selected for templates. We argue that sentence generation can be driven only through semantic and proper selection of the semantic layer composition and can enhance the accurate description of the images.

In this work, we have mainly concentrated on the viability of the other language-based image captioning systems. While English is a structural language (syntactic and grammatical), there are unstructured languages (only grammatically) like most of the Indian languages, which are derived from Sanskrit like Bengali (90\% Sanskrit vocabulary), Tamil (70\% Sanskrit vocabulary), Hindi (50\% Sanskrit vocabulary), etc. to name a few. We will be mainly focusing on the Bengali language as a case study and demonstrate how they do for the image captioning counterparts. We have used the Google Translation API (googletrans.Translator) for our work. Though the Bengali language is unstructured, the captioning model performed well for generating the captions. However, this is the first of its kind, never tried research-topic. Due to the absence of any such dataset, we have used a translation application for data generation. It is justified that such a model can be built for a language like Bengali, as it is the fifth-largest widely spoken language in the world and is one of the primitive languages of the world. We have also demonstrated that if we provide a Gaussian smoothening of the semantic layer, we can deliver better captions.

\begin{figure*}[!h] 
\centering 
\includegraphics[width=.8\textwidth]{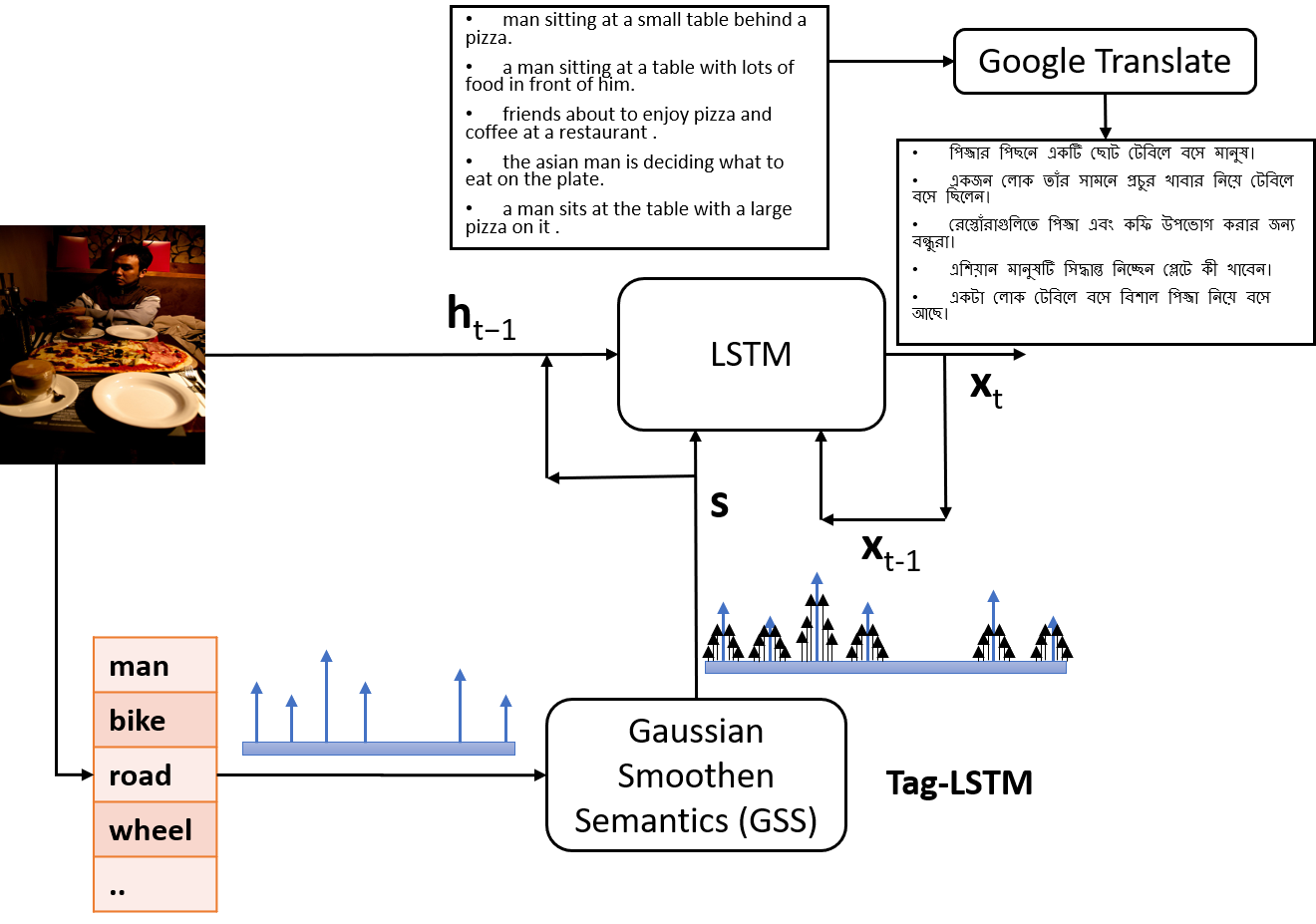} 
\caption{Overview of The Image Description Generation Methodology Using Google Translation (Google Translation API (googletrans.Translator)).}
\label{fig:overview}
\end{figure*}

\begin{figure*}[!h] 
\centering 
\includegraphics[width=.9\textwidth]{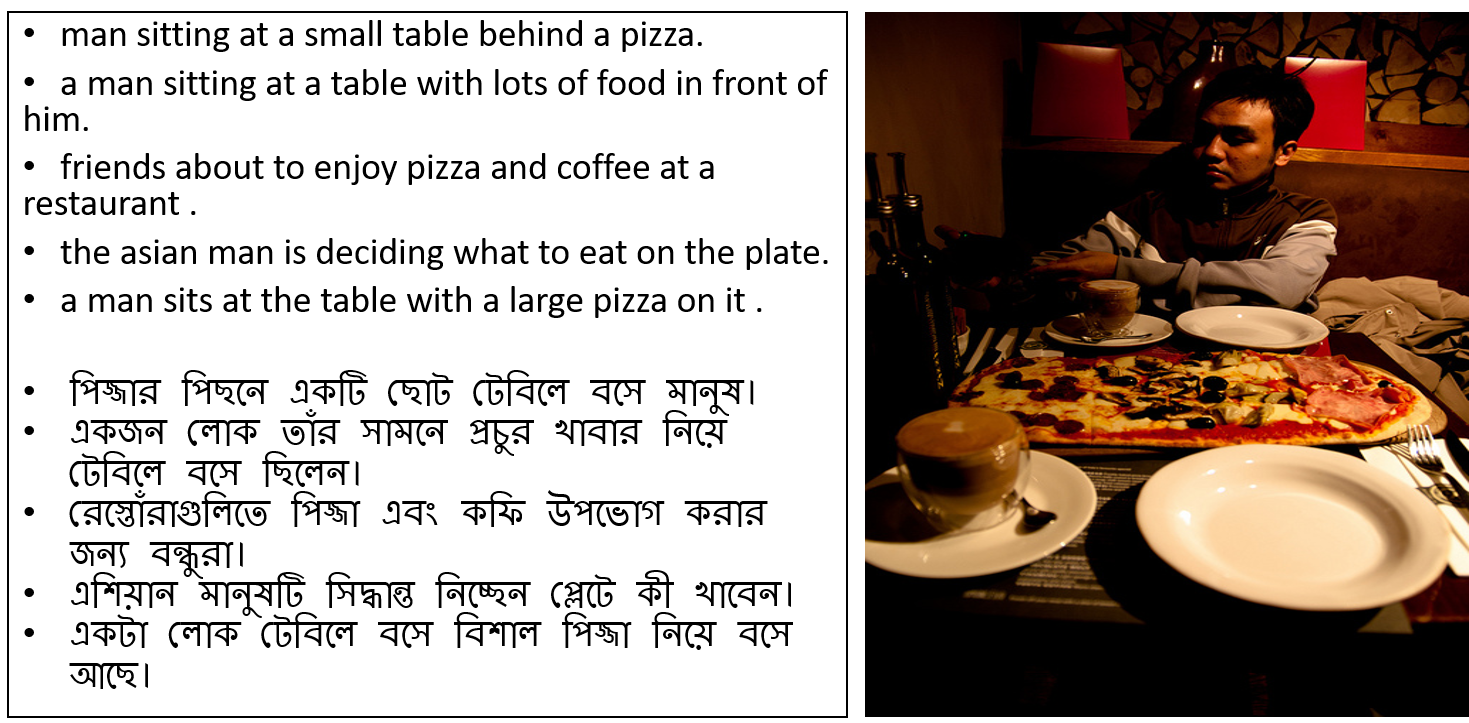} 
\caption{Example Instances of Generated Translated Data Along with English References}
\label{fig:instance}
\end{figure*}

\subsection{Image Caption Problem} 
The problem of image caption \cite{sur2019survey} is rooted in understanding the network, relationship, interaction \cite{Krishna2017} of the different objects and descriptions. These connections are evident to humans but it had to be taught to machines to be able to leverage from these visual features. It is difficult to generalize for different images if the sentences are generated from the object combinations through the determination of the sequentiality and priority of these objects, it is an NP-Hard problem. The models are made to determine the region and the objects in the image heuristically and then the description. It is said to be more than object detection and object subcategory or details interpretation. 
Image captioning in regional languages require immense exploration with the increasing number of applications expressing with images. Also, a large part of the Indian subcontinent is still dependent on regional language as their source of expression. These regional language-based expressions comprise a form of art as these applications integrate well and get constantly improved through human intervention.
Image captioning in different Indian languages has its challenges, where each of these languages possesses a different set of language attributes, grammatical properties, and completely different from English and other European languages. These Indian languages are entirely different from their root language Sanskrit. The diversification of these languages was created through different oral transformations, which was evident in ancient India due to the absence of texts and learning centers for everyone. Languages and customs propagated through ``Smriti" (or oral propagation) through different people, which later transferred to other parts of the world, including Europe. This fact is evident from the common words like ``Matri'' became Mother, ``Pitri" became Father, ``Brahta'' became Brother, ``Jamity" become Geometry and even ``Om" became Amen. However, from the prospect of grammar and way of construction, Indian languages are far different from Latin-originated languages, and in this work, we made an effort to generate sentences in these primitive and ancient culture languages.

The rest of the document is arranged with revisit of the existing works in literature in Section \ref{section:literature},
the description and statistics of the data in Section \ref{section:data},
learning network and representation detailed description in Section \ref{section:learning},
the intricacies of our methodology in Section \ref{section:methodology}, 
experiments, results and analysis in Section \ref{section:results},
concluding remarks with future prospects in Section \ref{section:discussion},

The main contribution of this work: 1) Bengali (Indian) Language Caption Generators 2) Framework for Translator based training 3) Gaussian Smoothen Semantic Features for Better Semantic Selection 4) Compositional-Decompositional Network Framework for better representation for captions.

\section{Reference Exploitation \& Google Translate} \label{section:data}
There are around 31 official languages, while the constitution recognizes 22 of these languages. As a result, every such language provides a different prospect of these images, and the development of machine generation in these languages provide better solutions to many applications, including automatic communicator, regional aid provider, and the list is endless. In this work, we made an effort to develop a caption generator and generated a training dataset of captions from the existing English MSCOCO dataset through the use of Google Translation API (googletrans.Translator). While it is true that all the translated captions are not perfect, but it can, at least, provide a good starting point and provide some evaluation standard.
\begin{table}[!h]
\centering
\caption{Statistics of Training Data for Different Indian Languages. Statistical Estimation of Vocabulary Size and Vocabulary Appearance Provides Insights into the Capability of Networks to Learn}
\begin{tabular}{|c|c|c|}
\hline
Language & Population & Vocabulary \\ 
\hline \hline
English & 8791 & 8K \\ \hline
Bengali & 9185 & 20K \\ \hline
\end{tabular}
\label{table:stats}
\end{table}
Table \ref{table:stats} provided an estimation of the vocabulary sizes, while Figure \ref{fig:instance} provided some instances of the translated captions in different languages for some images. However, we have considered some part of the vocabulary, say top 50\%, for likelihood generation, as the whole vocabulary is considerably long, and there are chances of sparsity and negligence of some words due to scarcity of samples in sentences. To find the correct set of vocabulary, we experimented to determine the best set and the cutoff. 
In Figure \ref{fig:instance}, being a native speaker of Bengali, we have provided other captions that would be much more realistic than the translated version.

\section{Learning Network and Representation Description} \label{section:learning}
Learning the representations and correct configuration of the network is essential for the success of language generators. While image feature is helpful, semantic distribution features help in enhancing the captions through direct correlation of the selected objects in images and the likelihood of words in sentences. In this work, we have introduced new architectures utilizing different feature spaces and reported their qualitative and quantitative analysis. 
While the semantic features are highly sparse, we introduced a Gaussian smoothing for semantic features and shown that the caption gets improved with these kinds of smoothing. Smoothing helps in better capturing of the image region as attention to be decomposed at the hidden level. 
In this work, we have focused on compositional and decomposition-based architectures as they provide some of the state-of-the-art performance \cite{Gan2016} for caption generator without the explicit requirement for large scale image segmentation and region-based feature extraction like \cite{anderson2018bottom}. However, they are good ways of expansion of the network and generate better descriptions for images. Our argument is that we can derive better network through decomposition of the causal representation $\textbf{h}_{t-1}$ than decomposition of both causal $\textbf{h}_{t-1}$ and previous $\textbf{x}_{t-1}$ contexts. Sequential recurrent neural network has been widely used in generative applications and mathematically, the semantic concept layer is initialized for these networks as the following,
 
\begin{equation} \label{eq:h0_mlp}
\textbf{h}_0 = MLP_h(\textbf{v})
\end{equation}
\begin{equation} \label{eq:c0_mlp}
\textbf{c}_0 = MLP_c(\textbf{v})
\end{equation}
where $\textbf{v}$ refers to visual features and $MLP(.)$ referred to multiple-layer perceptron. Figure \ref{fig:overview} provided a generalized architectural overview of the image caption generation system. Later these layer is fused with the Gaussian smoothen semantic features and provided with the perfect context representation for each word. 
\begin{figure*}[!h] 
\centering 
\includegraphics[width=\textwidth]{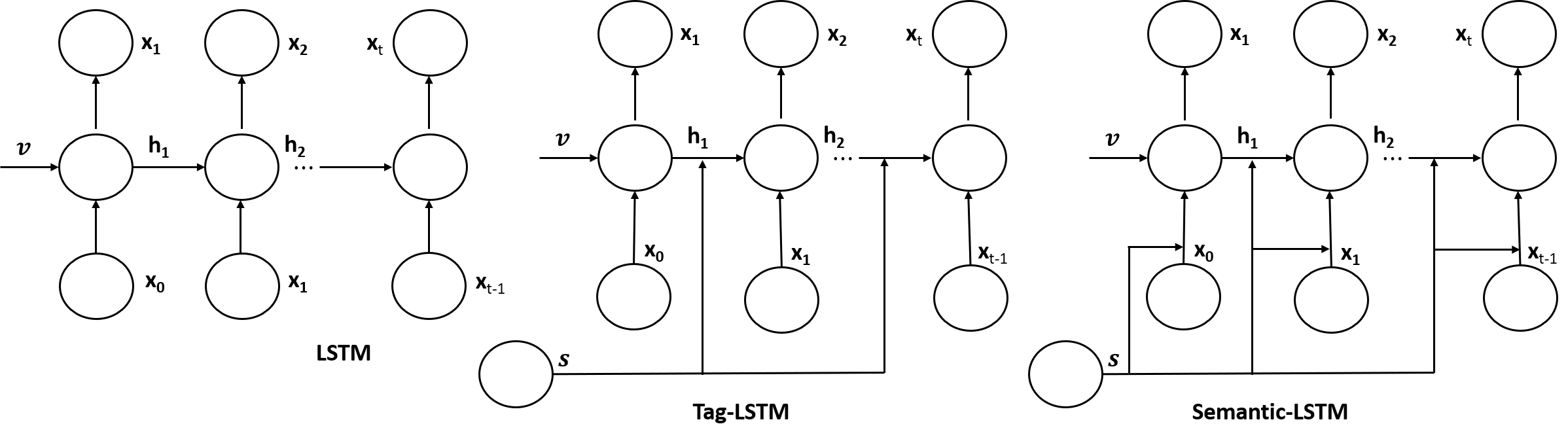} 
\caption{Architectural Comparison of Different LSTM Types: LSTM, GSTag-LSTM and GSSCN-LSTM.}
\label{fig:comparison}
\end{figure*}

\subsection{Long Short Term Memory} 
Long Short Term Memory (LSTM) is widely used sequential recurrent unit because of its scalability and its unmatched generalized solutions for many sequential generation related to languages and other distinct representation learning for classification and most importantly, apart from suppression of the variations or approximation, like any other deep learning architecture, it helps in better generalization. LSTM caption generator, with ResNet features, performs best among all single context processing units for image captioning. The basic equations for LSTM is provided as the following set of equations. 
\begin{equation}
\textbf{i}_{t} = \sigma(\textbf{W}_{xi}\textbf{x}_{t-1} + \textbf{W}_{hi}\textbf{h}_{t-1} + \textbf{b}_{i})
\end{equation}
\begin{equation}
\textbf{f}_{t} = \sigma(\textbf{W}_{xf}\textbf{x}_{t-1} + \textbf{W}_{hf}\textbf{h}_{t-1} + \textbf{b}_{f})
\end{equation}
\begin{equation}
\textbf{o}_{t} = \sigma(\textbf{W}_{xo}\textbf{x}_{t-1} + \textbf{W}_{ho}\textbf{h}_{t-1} + \textbf{b}_{o})
\end{equation}
\begin{equation}
\textbf{g}_{t} = \tanh(\textbf{W}_{xg}\textbf{x}_{t-1} + \textbf{W}_{hg}\textbf{h}_{t-1} + \textbf{b}_{g})
\end{equation}
\begin{equation}
\textbf{c}_{t} = \textbf{f}_{t} \odot \textbf{c}_{t-1} + \textbf{i}_{t} \odot \textbf{g}_{t} 
\end{equation}
\begin{equation}
\textbf{h}_{t} = \textbf{o}_{t} \odot \tanh(\textbf{c}_{t})
\end{equation}
\begin{equation} \label{eq:yt}
\textbf{y}_{t} = \textbf{h}_t \textbf{W}_{hy}
\end{equation}
For categorization, $\textbf{y}_{t}$ is evaluated through convergence to the categorization distribution through softmax layer defined as, 
\begin{equation} \label{eq:yt_softmax}
\textbf{y}_{t} = \sigma(\textbf{y}_{t}) =\frac{\exp(\textbf{y}_{t})}{\sum\limits^{C}_{k=1} \exp((y_{t})_{k})} 
\end{equation}
where we have $\sigma(\textbf{y}_{t}) \in \mathbb{R}^{C} \in [0,1]^{C}$ with $C$ as the set of categories. Also we have $\textbf{x}_t \in \mathbb{R}^m$, $\textbf{y}_{t} \in \mathbb{R}^{C}$, $\textbf{W}_{hy} \in \mathbb{R}^{C\times d}$, $\textbf{i},\textbf{f},\textbf{o},\textbf{g},\textbf{c} \in \mathbb{R}^{d}$, $\textbf{h}_t \in \mathbb{R}^d$, $\textbf{W}_{x*} \in \mathbb{R}^{d\times m}$, $\textbf{W}_{h*},\textbf{W}_{c*} \in \mathbb{R}^{d\times d}$, $\textbf{b}_{*} \in \mathbb{R}^d$.
The objective minimization function is defined as $J(\textbf{W}) = \arg \min \frac{1}{2s} \sum \limits_{\forall s} \sum \limits_{i} ||y_{t,i} - y_{t,i}' ||^2 = \arg \min \frac{1}{2s} \sum \limits_{\forall s} ||\textbf{y}_t - \textbf{y}_t' ||^2 $ and $s$ number of training samples. The parameters are updated with $\alpha \frac{\partial J(\textbf{W})}{\partial \textbf{W}}$. Value of $\alpha$ determines the learning rate for adaption with the changing topology of the objective function space.
\begin{equation} \label{eq:xt}
\textbf{x}_{t} = (\arg \max (\textbf{h}_t \textbf{W}_{hy}) ) \textbf{W}_{E} = (\arg \max \textbf{y}_{t}) \textbf{W}_{E}
\end{equation}
where $\textbf{W}_{E} \in \mathbb{R}^{V \times d}$ is the word embedding representation of vocabulary size $V$ and each word is represented with vector of dimension $d$. In most of the architecture, $\textbf{h}_{t} \rightarrow \textbf{x}_{t} $ involves Equation \ref{eq:xt} and is a convenience and may not be explicitly mentioned with every model but is implied. Extra set of improvement is 

Mathematically, Long Short Term Memory (LSTM) model, denoted as $f_{_{L}}(.)$, can be described as the followings probability distribution estimation.
\begin{equation} \label{eq:lstm}
\begin{split}
f & _{_{L}}(\textbf{v}) = \prod\limits_{k}^{} \mathrm{Pr}(w_k \mid \textbf{v}, \text{ } \textbf{W}_{L_1}) \prod\limits_{}^{} \mathrm{Pr}(\textbf{v} \mid I, \text{ }\textbf{W}_1) \\ 
& = \prod\limits_{k}^{} \mathrm{Q}_{IC}(w_k \mid \textbf{v}) \prod\limits_{}^{} \mathrm{Q}(\textbf{v} \mid I) 
\end{split} 
\end{equation}
using the weights of the LSTM in the architecture is denoted as $\textbf{W}_{L_1}$, $w_i$ as words of sentences, $\textbf{v}$ as image features, $\mathrm{Q}_{IC}(.)$ and $\mathrm{Q}(.)$ are the Image Caption and feature generator function respectively. $\mathrm{Q}(.)$ derives $\textbf{v}$ from image $\textbf{I}$.

\subsection{Gaussian Smoothened Tag Long Short Term Memory} 
Gaussian Smoothened Tag Long Short Term Memory (GST-LSTM) involves consecutive operations of selection and compositional fusion of different context features, including a semantic layer feature set for better representation. Unlike GSSCN-LSTM, where the input and hidden layer are decomposed through fusion, GST-LSTM only involves decomposition of the hidden layer. The hidden layer gets fade away very quickly, and through decomposition, it gets regenerated and revived. GST-LSTM involves decomposition of the hidden layer transformation weights as $\textbf{W} = \textbf{W}_s\textbf{s}\textbf{W}_p$ where $\textbf{s}$ is directly involved as refined semantic level information and diagonally sparse for orthogonal selection. Through this architecture, we demonstrate that despite low weights, it can perform as competitively (or even better) as GSSCN-LSTM. In principle, a lower number of weights and properly justified data fusion can help in better inference. 
GST-LSTM architecture is marked by the reduction in weights compared to GSSCN-LSTM and performs equally well as GSSCN-LSTM for image caption generations. For a model with 512 dimensions, GST-LSTM has 20\% fewer weights than GSSCN-LSTM, but the reduction is directly proportional to the dimension of $\textbf{s}$. 
GST-LSTM Network with image features involves the same extra set of equations, where the the initial states $\textbf{h}_0$, $\textbf{c}_0$ are derived from Equation \ref{eq:h0_mlp} and Equation \ref{eq:c0_mlp} respectively. However, after each iteration, the hidden state gets revised by Equation \ref{eq:tag} through the selection of sub-region with the tag features $S$ before the next prediction. The initial state for image memory of hidden states $\textbf{h}_t$, $\textbf{c}_t$ get fades away with time, and extra attention can revive the image quality of the memory and enhance the performance. The main iteration of the model consists of the following set of equations. 
\begin{equation}
\hat{\textbf{S}} = f_{{s}_m}(\textbf{S})
\end{equation}
where we have defined $f_{{s}_m}(.)$ as the Gaussian Smoothing function. $\textbf{S}$ is sparse and operates as a diagonal matrix. In the diagonal matrix, the columns are orthogonal to each other, and the task of each column is to select the feature of interest and discard the others. These columns come directly from the image, and hence we talk about a column tensor and the whole matrix. As we operate different transformations and train the weights, the strength of these individual columns degrades or distributed unevenly. Hence, Gaussian smoothing will help in absorbing better features as it transforms the discrete points to a better structural absorbent. We call $\hat{\textbf{S}}$ as Gaussian Smoothen Semantic Features (GSSF). 
\begin{figure}
\centering 
\includegraphics[width=.5\textwidth]{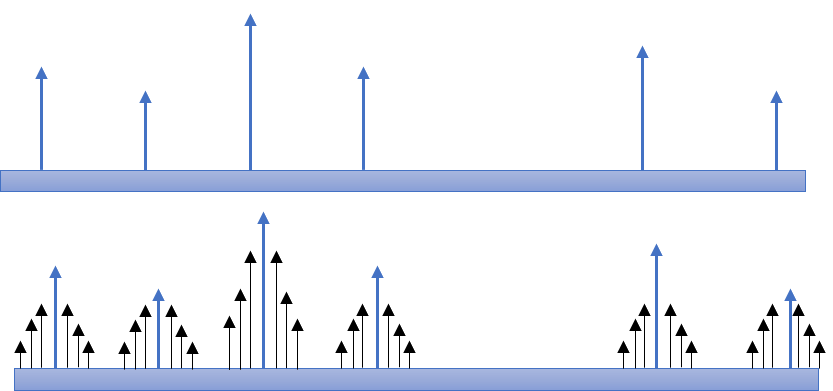} 
\caption{Explanation of Gaussian Smoothing of Semantic Features}
\label{fig:gauss}
\end{figure}
Due to limitations in GPU memory, we have kept the dimension of $\textbf{S}$ and $\hat{\textbf{S}}$ same.
\begin{equation} \label{eq:tag}
\textbf{h}_{t-1} = \textbf{W}_{h,m} \hat{\textbf{S}} \odot \textbf{W}_{h,n} \textbf{h}_{t-1}
\end{equation}
\begin{equation}
\textbf{i}_{t} = \sigma(\textbf{W}_{xi}\textbf{x}_{t-1} + \textbf{W}_{hi}\textbf{h}_{t-1} + \textbf{b}_{i})
\end{equation}
\begin{equation}
\textbf{f}_{t} = \sigma(\textbf{W}_{xf}\textbf{x}_{t-1} + \textbf{W}_{hf}\textbf{h}_{t-1} + \textbf{b}_{f})
\end{equation}
\begin{equation}
\textbf{o}_{t} = \sigma(\textbf{W}_{xo}\textbf{x}_{t-1} + \textbf{W}_{ho}\textbf{h}_{t-1} + \textbf{b}_{o})
\end{equation}
\begin{equation}
\textbf{g}_{t} = \tanh(\textbf{W}_{xg}\textbf{x}_{t-1} + \textbf{W}_{hg}\textbf{h}_{t-1} + \textbf{b}_{g})
\end{equation}
\begin{equation}
\textbf{c}_{t} = \textbf{f}_{t} \odot \textbf{c}_{t-1} + \textbf{i}_{t} \odot \textbf{g}_{t} 
\end{equation}
\begin{equation}
\textbf{h}_{t} = \textbf{o}_{t} \odot \tanh(\textbf{c}_{t})
\end{equation}

\begin{figure*}[!h] 
\centering 
\includegraphics[width=\textwidth]{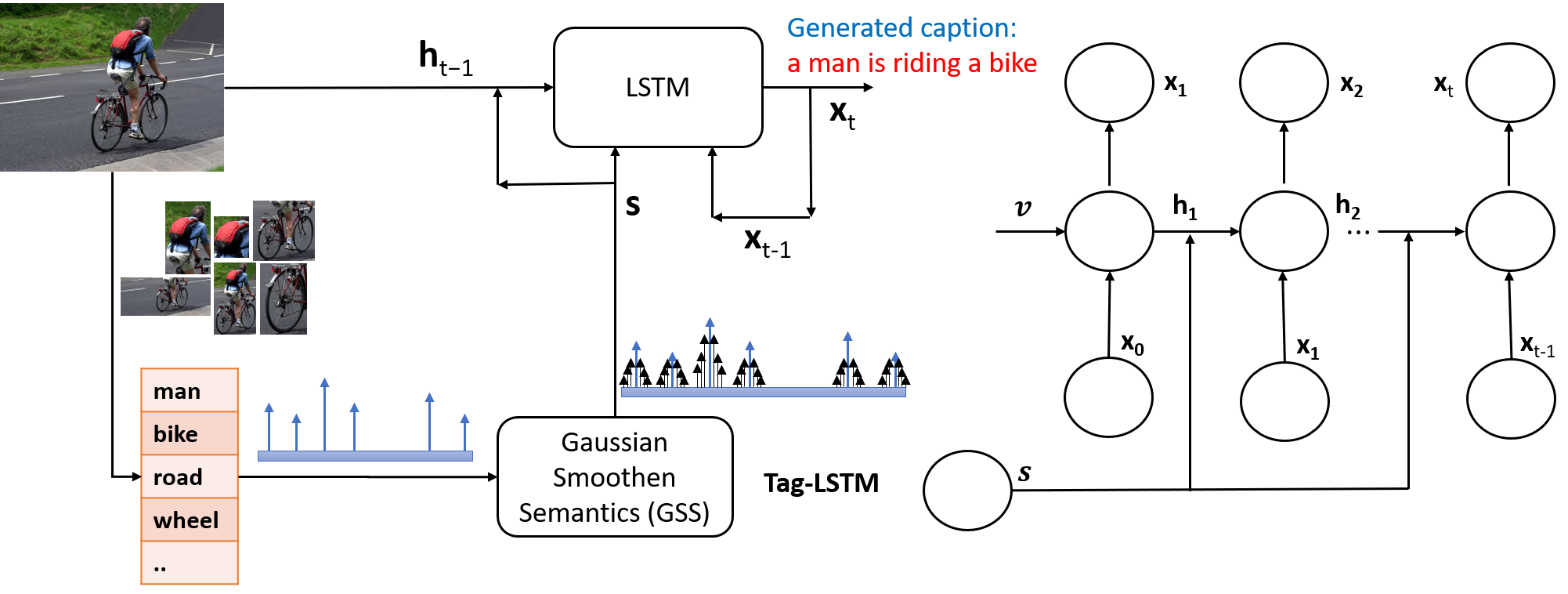} 
\caption{Overall Architecture of GST-LSTM Caption Generator with Gaussian Smoothen Semantic Features.}
\label{fig:GaussianSmoothing_Tag}
\end{figure*}

Mathematically, Gaussian Smoothened Tag Long Short Term Memory (GST-LSTM), denoted as $f_{_{GST}}(.)$, can be described as the followings probability distribution estimation.
\begin{equation} \label{eq:gst}
\begin{split}
f & _{_{GST}}(\textbf{v}) = \prod\limits_{k}^{} \mathrm{Pr}(w_k \mid \textbf{v}, \text{ } \hat{\textbf{S}} \odot \textbf{h}_{t-1}, \text{ } \textbf{W}_{L_1}) \prod\limits_{}^{} \mathrm{Pr} (\textbf{S} \mid \textbf{v}, \text{ }\textbf{W}_2) \\
& \prod\limits_{}^{} \mathrm{Pr}(\textbf{v} \mid I, \text{ }\textbf{W}_1) \\ 
& = \prod\limits_{k}^{} \mathrm{Q}_{IC}(w_k \mid \textbf{v}, \text{ } \hat{\textbf{S}}\odot \textbf{h}_{t-1}) \prod\limits_{}^{} \mathrm{Q}_S(\textbf{S} \mid \textbf{v}) \prod\limits_{}^{} \mathrm{Q}(\textbf{v} \mid I) 
\end{split} 
\end{equation}
using the weights of the LSTM in the architecture is denoted as $\textbf{W}_{L_1}$, $w_i$ as words of sentences, $\textbf{v}$ as image features, $\textbf{S}$ as semantic features from $\mathrm{Q}_S(.)$, $\hat{\textbf{S}}$ as Gaussian Smoothened features, $\mathrm{Q}_{IC}(.)$ and $\mathrm{Q}(.)$ are the Image Caption and feature generator function respectively. $\mathrm{Q}(.)$ derives $\textbf{v}$ from image $\textbf{I}$.

\subsection{Gaussian Smoothened Semantic Long Short Term Memory} 
Gaussian Smoothened Semantic Long Short Term Memory (GSSCN-LSTM) operated a combination of decomposition of both previous context and hidden layer information and compared to GST-LSTM, it involves further decomposition of features, which are fused to compose representations at a higher level for caption generation, but at the cost of series of weighted transformation weights. The set of equations that describe GSSCN-LSTM is provided below. (\cite{kiros2014multimodal,Gan2016}) used the decomposition techniques in their architecture. 
\begin{equation}
\hat{\textbf{S}} = f_{{s}_m}(\textbf{S})
\end{equation}
where we have defined $f_{{s}_m}(.)$ as the Gaussian Smoothing function. 
\begin{equation}
\textbf{x}_{*,t} = \textbf{W}_{x,*m} \hat{\textbf{S}} \odot \textbf{W}_{x,*n} \textbf{x}_{t-1}
\end{equation}
\begin{equation}
\textbf{h}_{*,t} = \textbf{W}_{h,*m} \hat{\textbf{S}} \odot \textbf{W}_{h,*n} \textbf{h}_{t-1}
\end{equation}
where we have ${\textbf{S}}$ as the tag distribution and $\hat{\textbf{S}}$ is the Gaussian smoothen tag distribution, $\textbf{v}$ as the image features, $* = i/f/o/g$, $\textbf{W}_{*m} \in \mathbb{R}^{a \times b}$, 
$\textbf{W}_{*n} \in \mathbb{R}^{a \times b}$, 
$\textbf{h}_{*,t} \in \mathbb{R}^{a \times b}$ and 
$\textbf{x}_{*,t} \in \mathbb{R}^{a \times b}$.
The decomposed features are processed in the memory network through the following set of equations. 
\begin{equation}
\textbf{i}_t = \sigma(\textbf{W}_{xi}\textbf{x}_{i,t} + \textbf{W}_{hi}\textbf{h}_{i,t-1} + \textbf{b}_{i})
\end{equation}
\begin{equation}
\textbf{f}_t = \sigma(\textbf{W}_{xf}\textbf{x}_{f,t} + \textbf{W}_{hf}\textbf{h}_{f,t-1} + \textbf{b}_{f})
\end{equation}
\begin{equation}
\textbf{g}_t = \sigma(\textbf{W}_{xg}\textbf{x}_{g,t} + \textbf{W}_{hg}\textbf{h}_{g,t-1} + \textbf{b}_{g})
\end{equation}
\begin{equation}
\textbf{o}_t = \sigma(\textbf{W}_{xo}\textbf{x}_{o,t} + \textbf{W}_{ho}\textbf{h}_{o,t-1} + \textbf{b}_{o})
\end{equation}
\begin{equation}
\textbf{c}_{t} = \textbf{f}_{t} \odot \textbf{c}_{t-1} + \textbf{i}_{t} \odot \textbf{g}_{t} 
\end{equation}
\begin{equation}
\textbf{h}_t = \textbf{o}_{t} \odot \tanh(\textbf{c}_{t})
\end{equation}
where $\textbf{W}_{x*} \in \mathbb{R}^{w_e \times s}$, $\textbf{W}_{h*} \in \mathbb{R}^{d \times s}$, where $d$ is the hidden layer dimension, $w_e$ is the word embedding dimension and $s$ is the semantic dimension. 
Here, we have replaced semantic feature $\hat{\textbf{S}}$ with Gaussian Smoothen Semantic Features (GSSF) $\hat{\textbf{S}}$ and a gradual smoothen feature helps in better selection for the next phase from the previous states, generated from the model. It must be mentioned that $\textbf{h}_t$ is shadow image representation generated from the image features and continuous fusion of the semantic features and the image features.

Mathematically, Gaussian Smoothened Semantic Long Short Term Memory (GSSCN-LSTM), denoted as $f_{_{GSSCN}}(.)$, can be described as the followings probability distribution estimation.
\begin{equation} \label{eq:gsscn}
\begin{split}
f & _{_{GSSCN}}(\textbf{v}) = \prod\limits_{k}^{} \mathrm{Pr}(w_k \mid \textbf{v}, \text{ } \hat{\textbf{S}} \odot \textbf{h}_{t-1}, \text{ } \hat{\textbf{S}}\odot \textbf{x}_{t-1}, \text{ } \textbf{W}_{L_1}) \\
& \prod\limits_{}^{} \mathrm{Pr} (\textbf{S} \mid \textbf{v}, \text{ }\textbf{W}_2) \prod\limits_{}^{} \mathrm{Pr}(\textbf{v} \mid I, \text{ }\textbf{W}_1) \\ 
& = \prod\limits_{k}^{} \mathrm{Q}_{IC}(w_k \mid \textbf{v}, \text{ } \hat{\textbf{S}}\odot \textbf{h}_{t-1}, \text{ } \hat{\textbf{S}}\odot \textbf{x}_{t-1}) \\
& \prod\limits_{}^{} \mathrm{Q}_S(\textbf{S} \mid \textbf{v}) \prod\limits_{}^{} \mathrm{Q}(\textbf{v} \mid I) 
\end{split} 
\end{equation}
using the weights of the LSTM in the architecture is denoted as $\textbf{W}_{L_1}$, $w_i$ as words of sentences, $\textbf{v}$ as image features, $\textbf{S}$ as semantic features through $\mathrm{Q}_S(.)$, $\hat{\textbf{S}}$ as Gaussian Smoothened features, $\mathrm{Q}_{IC}(.)$ and $\mathrm{Q}(.)$ are the Image Caption and feature generator function respectively. $\mathrm{Q}(.)$ derives $\textbf{v}$ from image $\textbf{I}$.

\begin{figure*}[!h] 
\centering 
\includegraphics[width=\textwidth]{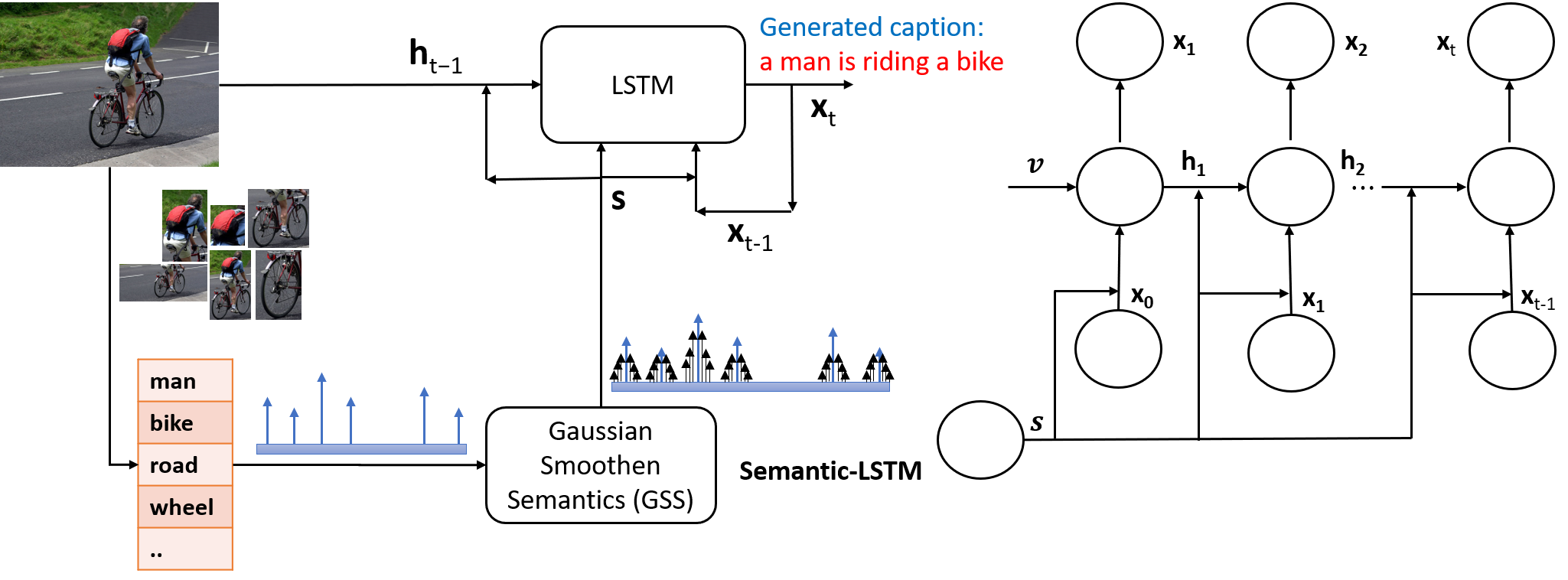} 
\caption{Overall Architecture of GSSCN-LSTM Caption Generator with Gaussian Smoothen Semantic Features.}
\label{fig:GaussianSmoothing_SCN}
\end{figure*}

\section{Details of Methodology} \label{section:methodology}
We have provided some implementation level information, though this is very common for language generator and image captioning research. There are intricacies of the procedures and are considered as the following. 
\paragraph{Step 1: Data Acquisition} We have used the MSCOCO data, and some of them are shared by \cite{Gan2016}, which coordinates with the ResNet features and can be easily obtained. 
\paragraph{Step 2: CNN Transfer Learning Feature} Here, we have extensively used ResNet through transfer learning with 2048 dimension. 
\paragraph{Step 3: Transfer Learning Translation} The Bengali captions are obtained through Google Translation API (googletrans.Translator) and a plugin provided by them. Since it is a public plugin and protected by DOS attack, the number of times it can translate for you on a certain day is limited, and hence it took me several days to translate the data. Also, since Bengali is unstructured language, the vocabulary dimension is very large and needs to be cleaned and processed with by 5\% occurrence rate. The English vocabulary is also needed to be shrink-ed by 5\% occurrence rate.
\paragraph{Step 4: Transfer Learning Semantic} After the CNN features are extracted, Tag features with 999 dimensions are used, and this is a neural network transformation of the image features.
\paragraph{Step 5: Gaussian Smoothing of Semantic} Here, we explain and introduced Gaussian Smoothen Semantic Features (GSSF). Figure \ref{fig:gauss} provided an instance for an illustration of the effects of the Gaussian Smoothing procedure on semantic features.
\paragraph{Step 6: Training Sessions} We trained the LSTM network end-to-end through the use of the all available data, but without the CNN network as it becomes hefty and prone to over-fitting. 
\paragraph{Step 7: Testing in Original Language Space} We processed the evaluation of the generated Bengali sentences with the ground-truth Bengali sentences. 
\paragraph{Step 8: Testing in Reference Language Space} For better analysis and clarity, we translated the Bengali sentences and evaluated the performances in the English language to get an overview in comparison to the English language caption generator. Also, it provides some reference of how far we need to go before we can get a perfect captioner. But, unlike English, which is a grammatically structured language, Bengali (with 90\% Sanskrit vocabulary) is an unstructured language, and direct comparison of relative performance is difficult. 
\paragraph{Step 9: Qualitative Evaluation for Semantics Correctness} Also, we have made some semantic relatedness through visual inspection in Qualitative Evaluation section.

\section{Results \& Analysis} \label{section:results}
This section provides some details of the experiments. We have used different evaluation techniques to put up different aspects of the generated captions in Bengali language. We also provided instances of the generated caption in 
Evaluation of this kind of translation based caption generator analysis is based on two fronts, mainly because no expert can legitimatize the correctness of all these languages, and in the absence of grammar, the semantics of these languages are difficult to define. However, we can offer qualitative evaluations for the working versions of the sentences in these languages. Nevertheless, we have provided both qualitative and quantitative metric based comparisons below. 

\subsection{Dataset}
We have used the same training-validation-testing split \cite{Karpathy2015Deep} of MSCOCO data that has been used in all the papers, which consists of 123287 train+validation images and 566747 train+validation sentences. Here, each image is associated with at least five sentences from a vocabulary of 8791 words. There are 5000 images (with 25010 sentences) for validation and 5000 images (with 25010 sentences) for testing. We used the same data split, as described in Karpathy's paper \cite{Karpathy2015Deep}. For our network, two sets of features are being used: one is ResNet features with 2048 dimension feature vector, and another is the semantic representation with feature vector of 999 dimensions and consisted of the likelihood probability of occurrence for the most appearing set of objects, attributes and interaction criteria as set of combinations of incidents for the dataset. 
For the translated data, we have restricted the vocabulary of the sentences to 20K and used 32 and 64 batch size based training sessions for each of the models. Also, we found that 1024 dimensions of the hidden layers provided the best performance, while the word embedding is loaded from Stanford GloVe 300 dimension pre-trained model. 

\subsection{Evaluation Metrics}
Different metrics like CIDEr-D, METEOR, ROUGE\_L, BLEU\_n, and SPICE are used for our evaluation. However, we think that the BLEU\_4 metric is the most sensible way of evaluation as this metric technique takes care of the continuity of object-attribute relationships and at the same time provides organizes the best possible descriptions. Also, BLEU\_4 evaluation considers several consecutive lineages based on nouns, adjectives, adverbs from the ground truth, which can never be judged through other criteria. 
We used Microsoft COCO dataset for our evaluation, and the results are reported based on Karpathy's division of training, validation, and testing set. 

\subsection{Evaluation Procedures}
The first evaluation is in Bengal vocabulary space and the performance is dependent on the vocabulary space of the training model. If we consider the function of the translation as $Tr(.)$ and the evaluator function as $f_Q(.)$, then we define the two different evaluation criteria as the followings. 
\begin{equation}
E_1 = f_Q(\textbf{G}_L, Tr(\textbf{R}_E)_L) 
\end{equation}
\begin{equation}
E_2 = f_Q(Tr(\textbf{G}_L)_E, \textbf{R}_E) 
\end{equation}
where $Tr(.)_L$ is the translation function to language $L$ and $\textbf{G}_L$ is the generated captions in language $L$ and $\textbf{R}_L$ is the reference caption in language $L$. We define evaluator function $f_Q(.)$ as, 
\begin{equation}
f_Q(.) \in \{\text{BLEU\_n, CIDEr, ROUGE\_L, METEOR, SPICE} \}
\end{equation} 
Also, The two reference languages frames denoted as $L$ (Any Language other than English) and $E$ (English). 
The evaluation criteria had varied ranges of metrics, and we evaluated the generated sentences with each of them due to the limitation of grammatical knowledge of different languages and the way the machine generates captions in different languages. Apart from the different evaluation metrics, we have also used Beam Search for generating longer sentences or at least sentences with a higher cumulative probability value.

\subsection{Quantitative Analysis}
We made a comparative study of the different architectures and compositional extraction of different representations. Quantitative analysis helps in understanding the capability of the model in objects based inference in sentences and its distance from the ground truth and also their ability to compose a different form of short sequences from image contexts. Table \ref{table:performanceEva1} provided a comparative overview of evaluation based on the language itself based on the reference of that language. In contrast, Table \ref{table:performanceEva2} provided the same metric evaluation when the generated captions are translated to English language, and a comparison is made based on the ground truth of the reference of the English language. From Table \ref{table:performanceEva1} it is evident that the captioner generator performed very well. Even though the source of data is a translator, which has not reached perfection and is just a prototype with errors. This limitation of the translation models is mainly due to the unstructured nature of the ancient languages like Bengali, which is directly related to Sanskrit and was there from pre-historic times. 
From Table \ref{table:performanceEva2}, we can say that our analysis is limited to constrained vocabulary (20K most appearing Bengali words) and given a better data, this approaches can be improved with precision and as a native Bengali speaker, this model can solve the problem of Bengali language based image captioner applications. We have also used beam search with beam size as 5 and also made sure that the repetition is prevented from appearing as it reduces the performance. Also, we have utilized a dropout rate of 0.5 for the embedding, image features, entry points of the language decoder and the exit points of the language decoders. 0.5 dropout rate provides ample scope for all the variables to be selected or rejected entirely and is very helpful to avoid over-fitting. 

\subsection{Discussion}
For analysis of the results, we found that our proposed model GST-LSTM model performed the best and is also much lower (in the number of weights) in comparison to GSSCN-LSTM and better than LSTM base model in performance. GST-LSTM out-performed GSSCN-LSTM and LSTM by 1\% and 4\% respectively for BLEU\_4 metric and better data will even improve the relative performances. The main reason why the GSSF model performed better has diverse explanations. The addition of the semantic level information and increasing the spread using Gaussian Smoothening (GSSF) has helped considerably. This approach can improve the results as it is providing a better concept layer for the generation of the representation for the captions. 
GSSF provides a new technique to upgrade the sparse semantic tensor information, mainly when it is acting as a diagonal matrix and its prime work is the selection of the structural information from the media contents. Other factors that helped in better captioning is through the extraction of the usable vocabulary space and also restricting the word embedding. We have used the word embedding of Stanford GloVe through the pre-trained model and gradually smoothened it. Also, for assigning the embedding vector, we have taken the help of a translator model to map the Bengali words with its English counterparts. Table \ref{table:performanceEva1} and Table \ref{table:performanceEva2} provided the numerical results.

\begin{table*}
\centering
\caption{Performance Evaluation: Original Generated Vs Original Ground Truth}
\begin{tabular}{|c|c|c|c|c|c|c|c|}
\hline
Algorithm & Language & CIDEr-D & Bleu\_4 & Bleu\_3 & Bleu\_2 & Bleu\_1 & ROUGE\_L \\ 
\hline \hline 
LSTM & Bengali & 0.93 & 0.31 & 0.42 & 0.55 & 0.70 & 0.53  \\ \hline
GST-LSTM & Bengali & \textbf{0.99} & \textbf{0.35} & \textbf{0.48} & \textbf{0.63} & \textbf{0.77} & \textbf{0.57}  \\ \hline 
GSSCN-LSTM & Bengali & 1.01 & 0.34 & 0.44 & 0.58 & 0.73 & 0.55  \\ \hline
\end{tabular}
\label{table:performanceEva1}
\end{table*}

\begin{table*}
\centering
\caption{Performance Evaluation: English Translated of Generated Vs English Ground Truth}
\begin{tabular}{|c|c|c|c|c|c|c|c|}
\hline
Algorithm & Language & CIDEr-D & Bleu\_4 & Bleu\_3 & Bleu\_2 & Bleu\_1 & ROUGE\_L  \\ 
\hline \hline
LSTM & Bengali & 0.62 & 0.24 & 0.34 & 0.46 & 0.58 & 0.39  \\ \hline
GST-LSTM & Bengali & \textbf{0.80} & \textbf{0.25} & \textbf{0.36} & \textbf{0.49} & \textbf{0.66} & \textbf{0.46}  \\ \hline
GSSCN-LSTM & Bengali & 0.78 & 0.25 & 0.35 & 0.49 & 0.66 & 0.46  \\ \hline
\end{tabular}
\label{table:performanceEva2}
\end{table*}

\subsection{Qualitative Analysis}
\begin{figure*}[!h] 
\centering 
\includegraphics[width=\textwidth]{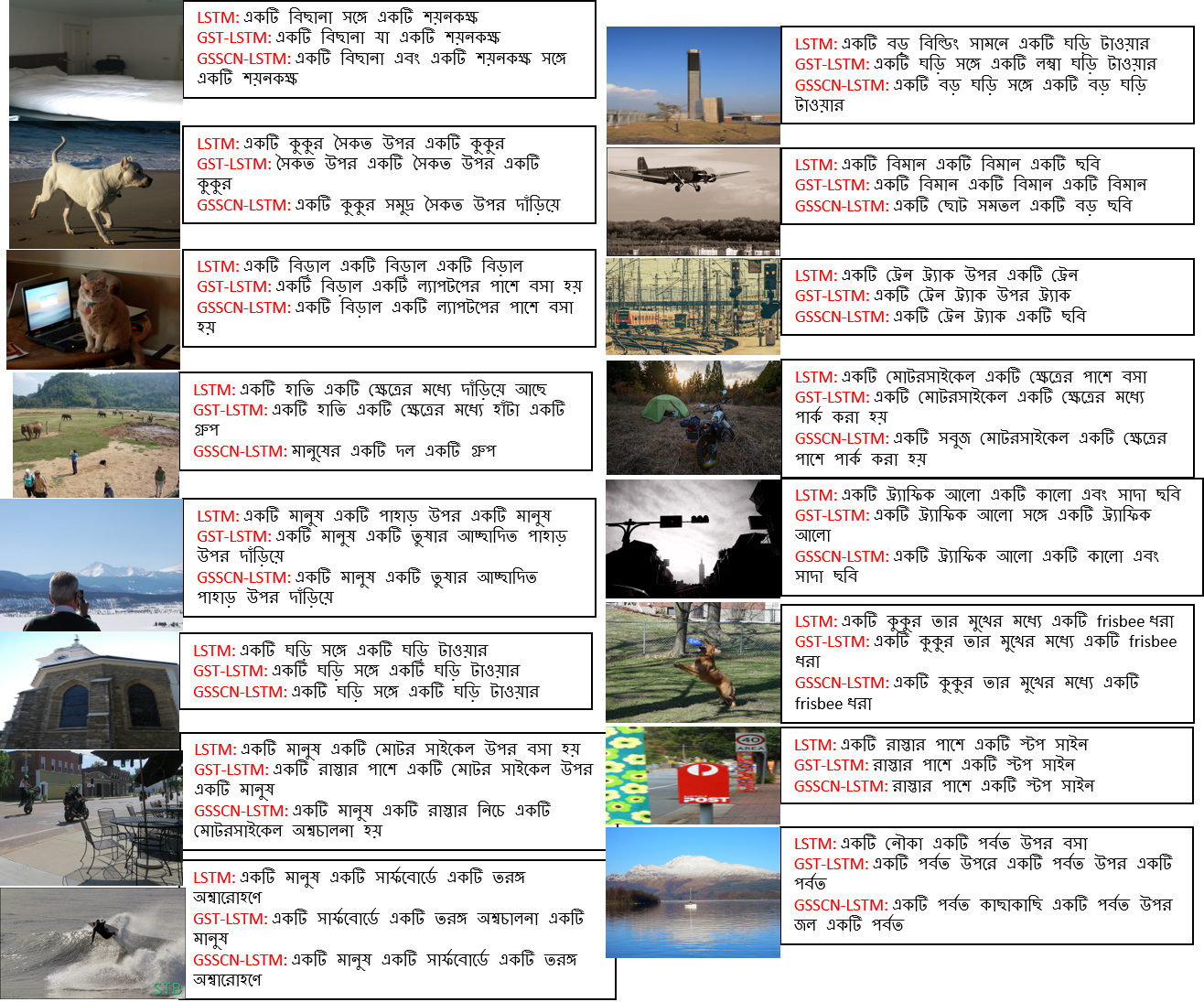} 
\caption{Qualitative Analysis of Captions. Set 1.}
\label{fig:qualitativeEvaluation1}
\end{figure*}

\begin{figure*}[!h] 
\centering 
\includegraphics[width=\textwidth]{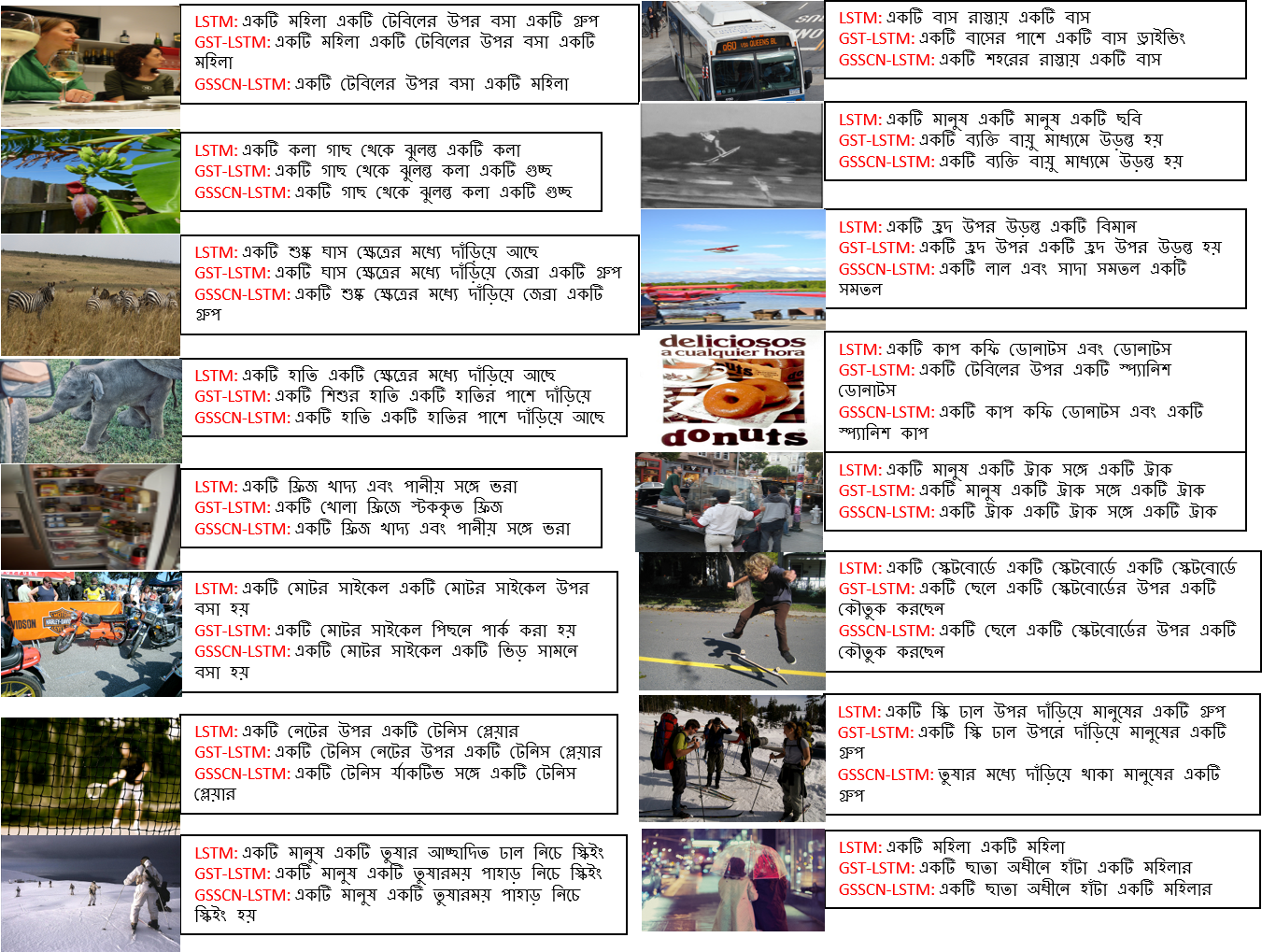} 
\caption{Qualitative Analysis of Captions. Set 2.}
\label{fig:qualitativeEvaluation2}
\end{figure*}

It is very difficult to evaluate whether a model is better than the other through averaged numerical. Hence, we have provided some instances of the generated captions in Figure \ref{fig:qualitativeEvaluation1} and Figure \ref{fig:qualitativeEvaluation2}. Whether overall improved captions are generated or not is also difficult to judge from numerical in Table \ref{table:performanceEva1} and Table \ref{table:performanceEva2}. Hence, the following qualitative analysis is adopted and will reflect some of them. Most of the caption generated work used diverse evaluation methods for the generated captions, but there is a requirement of analysis of quality and context correctness of these languages. 
Also, qualitative metrics are the best way of understanding the acceptability of the generated sentences and hence in this part, we have demonstrated some instances with the original images. Figure \ref{fig:qualitativeEvaluation1} and Figure \ref{fig:qualitativeEvaluation2} provided some instances of the different caption of different Indian languages. 
From these generated captions, we can say that the model performed quite well and the sentences represent much better description of the situations.

\section{Conclusion \& Future Works} \label{section:discussion}
In this work, we have introduced a new memory unit architecture for sequential learning and utilized it for image captioning applications for different unstructured (grammatically) Indian languages like Bengali and is the fifth widely spoken language in the world. The features include a different set of language attributes and grammatical properties and completely different from English. While we have used Bengali language as our reference, other Indian languages can be used in this way as well. The quantitative evaluation criteria for performance can be regarded as a reference, while the qualitative evaluation criteria can be seen as a perfect judge for many situation narrations for images. We devised different architectures (GST-LSTM and GSSCN-LSTM) to provide a comparison with baseline LSTM and also provided a new technique to upgrade the sparse semantic tensor information, mainly when it is acting as a diagonal matrix. Its prime work is a selection of the structural information from the media contents. Since this is the very first work, lots of further work can be done in this paradigm and our performance metric can be used for reference and baseline.

\section*{Acknowledgment}
The author has used University of Florida HiperGator, equipped with NVIDIA Tesla K80 GPU, extensively for the experiments. The author acknowledges University of Florida Research Computing for providing computational resources and support that have contributed to the research results reported in this publication. URL: http://researchcomputing.ufl.edu

\ifCLASSOPTIONcaptionsoff
  \newpage
\fi

%









\begin{thebibliography}{1}

\bibitem{sur2019survey}
Sur, C. (2019). Survey of deep learning and architectures for visual captioning-transitioning between media and natural languages. Multimedia Tools and Applications, 1-51.

\bibitem{lu2018entity}
Lu, D., Whitehead, S., Huang, L., Ji, H., \& Chang, S. F. (2018). Entity-aware Image Caption Generation. arXiv preprint arXiv:1804.07889.

\bibitem{lu2018neural} 
Lu, J., Yang, J., Batra, D., \& Parikh, D. (2018, March). Neural Baby Talk. In Proceedings of the IEEE Conference on Computer Vision and Pattern Recognition (pp. 7219-7228).

\bibitem{you2018image} You, Q., Jin, H., \& Luo, J. (2018). Image Captioning at Will: A Versatile Scheme for Effectively Injecting Sentiments into Image Descriptions. arXiv preprint arXiv:1801.10121.

\bibitem{melnyk2018improved}
Melnyk, I., Sercu, T., Dognin, P. L., Ross, J., \& Mroueh, Y. (2018). Improved Image Captioning with Adversarial Semantic Alignment. arXiv preprint arXiv:1805.00063.

\bibitem{wu2018joint}
Wu, J., Hu, Z., \& Mooney, R. J. (2018). Joint Image Captioning and Question Answering. arXiv preprint arXiv:1805.08389.

\bibitem{kilickaya2017data}
Kilickaya, M., Akkus, B. K., Cakici, R., Erdem, A., Erdem, E., \& Ikizler-Cinbis, N. (2017). Data-driven image captioning via salient region discovery. IET Computer Vision, 11(6), 398-406.

\bibitem{chen2017structcap}
Chen, F., Ji, R., Su, J., Wu, Y., \& Wu, Y. (2017, October). Structcap: Structured semantic embedding for image captioning. In Proceedings of the 2017 ACM on Multimedia Conference (pp. 46-54). ACM.

\bibitem{jiang2018learning}
Jiang, W., Ma, L., Chen, X., Zhang, H., \& Liu, W. (2018). Learning to guide decoding for image captioning. arXiv preprint arXiv:1804.00887.

\bibitem{wu2018modeling}
Wu, C., Wei, Y., Chu, X., Su, F., \& Wang, L. (2018). Modeling visual and word-conditional semantic attention for image captioning. Signal Processing: Image Communication.

\bibitem{fu2018image}
Fu, K., Li, J., Jin, J., \& Zhang, C. (2018). Image-Text Surgery: Efficient Concept Learning in Image Captioning by Generating Pseudopairs. IEEE Transactions on Neural Networks and Learning Systems, (99), 1-12.

\bibitem{chen2018show}
Chen, H., Ding, G., Lin, Z., Zhao, S., \& Han, J. (2018). Show, Observe and Tell: Attribute-driven Attention Model for Image Captioning. In IJCAI (pp. 606-612).

\bibitem{cornia2018paying}
Cornia, M., Baraldi, L., Serra, G., \& Cucchiara, R. (2018). Paying More Attention to Saliency: Image Captioning with Saliency and Context Attention. ACM Transactions on Multimedia Computing, Communications, and Applications (TOMM), 14(2), 48.

\bibitem{zhao2018multi}
Zhao, W., Wang, B., Ye, J., Yang, M., Zhao, Z., Luo, R., \& Qiao, Y. (2018). A Multi-task Learning Approach for Image Captioning. In IJCAI (pp. 1205-1211).

\bibitem{li2018coco}
Li, X., Wang, X., Xu, C., Lan, W., Wei, Q., Yang, G., \& Xu, J. (2018). COCO-CN for Cross-Lingual Image Tagging, Captioning and Retrieval. arXiv preprint arXiv:1805.08661.

\bibitem{chen2017reference}
Chen, M., Ding, G., Zhao, S., Chen, H., Liu, Q., \& Han, J. (2017, February). Reference Based LSTM for Image Captioning. In AAAI (pp. 3981-3987).

\bibitem{tavakoliy2017paying}
Tavakoliy, H. R., Shetty, R., Borji, A., \& Laaksonen, J. (2017, October). Paying attention to descriptions generated by image captioning models. In Computer Vision (ICCV), 2017 IEEE International Conference on (pp. 2506-2515). IEEE.

\bibitem{chen2017show}
Chen, H., Zhang, H., Chen, P. Y., Yi, J., \& Hsieh, C. J. (2017). Show-and-fool: Crafting adversarial examples for neural image captioning. arXiv preprint arXiv:1712.02051.

\bibitem{ye2018attentive}
Ye, S., Liu, N., \& Han, J. (2018). Attentive Linear Transformation for Image Captioning. IEEE Transactions on Image Processing.

\bibitem{wang2017skeleton}
Wang, Y., Lin, Z., Shen, X., Cohen, S., \& Cottrell, G. W. (2017). Skeleton key: Image captioning by skeleton-attribute decomposition. arXiv preprint arXiv:1704.06972.

\bibitem{chen2018factual}
Chen, T., Zhang, Z., You, Q., Fang, C., Wang, Z., Jin, H., \& Luo, J. (2018). " Factual" or" Emotional": Stylized Image Captioning with Adaptive Learning and Attention. arXiv preprint arXiv:1807.03871.

\bibitem{chen2018groupcap} 
Chen, F., Ji, R., Sun, X., Wu, Y., \& Su, J. (2018). GroupCap: Group-Based Image Captioning With Structured Relevance and Diversity Constraints. In Proceedings of the IEEE Conference on Computer Vision and Pattern Recognition (pp. 1345-1353).

\bibitem{liu2017mat}
Liu, C., Sun, F., Wang, C., Wang, F., \& Yuille, A. (2017). MAT: A multimodal attentive translator for image captioning. arXiv preprint arXiv:1702.05658.

\bibitem{harzig2018multimodal}
Harzig, P., Brehm, S., Lienhart, R., Kaiser, C., \& Schallner, R. (2018). Multimodal Image Captioning for Marketing Analysis. arXiv preprint arXiv:1802.01958.

\bibitem{liu2018show}
Liu, X., Li, H., Shao, J., Chen, D., \& Wang, X. (2018). Show, Tell and Discriminate: Image Captioning by Self-retrieval with Partially Labeled Data. arXiv preprint arXiv:1803.08314.

\bibitem{chunseong2017attend}
Chunseong Park, C., Kim, B., \& Kim, G. (2017). Attend to you: Personalized image captioning with context sequence memory networks. In Proceedings of the IEEE Conference on Computer Vision and Pattern Recognition (pp. 895-903).

\bibitem{sharma2018conceptual} 
Sharma, P., Ding, N., Goodman, S., \& Soricut, R. (2018). Conceptual Captions: A Cleaned, Hypernymed, Image Alt-text Dataset For Automatic Image Captioning. In Proceedings of the 56th Annual Meeting of the Association for Computational Linguistics (Volume 1: Long Papers) (Vol. 1, pp. 2556-2565).

\bibitem{yao2017incorporating}
Yao, T., Pan, Y., Li, Y., \& Mei, T. (2017, July). Incorporating copying mechanism in image captioning for learning novel objects. In 2017 IEEE Conference on Computer Vision and Pattern Recognition (CVPR) (pp. 5263-5271). IEEE.

\bibitem{zhang2017actor}
Zhang, L., Sung, F., Liu, F., Xiang, T., Gong, S., Yang, Y., \& Hospedales, T. M. (2017). Actor-critic sequence training for image captioning. arXiv preprint arXiv:1706.09601.

\bibitem{fu2017aligning}
Fu, K., Jin, J., Cui, R., Sha, F., \& Zhang, C. (2017). Aligning where to see and what to tell: Image captioning with region-based attention and scene-specific contexts. IEEE transactions on pattern analysis and machine intelligence, 39(12), 2321-2334.

\bibitem{ren2017deep} 
Ren, Z., Wang, X., Zhang, N., Lv, X., \& Li, L. J. (2017). Deep reinforcement learning-based image captioning with embedding reward. arXiv preprint arXiv:1704.03899.

\bibitem{liu2017improved}
Liu, S., Zhu, Z., Ye, N., Guadarrama, S., \& Murphy, K. (2017, October). Improved image captioning via policy gradient optimization of spider. In Proc. IEEE Int. Conf. Comp. Vis (Vol. 3, p. 3).

\bibitem{cohn2018pragmatically} 
Cohn-Gordon, R., Goodman, N., \& Potts, C. (2018). Pragmatically Informative Image Captioning with Character-Level Reference. arXiv preprint arXiv:1804.05417.

\bibitem{liu2017attention}
Liu, C., Mao, J., Sha, F., \& Yuille, A. L. (2017, February). Attention Correctness in Neural Image Captioning. In AAAI (pp. 4176-4182).

\bibitem{yao2017boosting}
Yao, T., Pan, Y., Li, Y., Qiu, Z., \& Mei, T. (2017, October). Boosting image captioning with attributes. In IEEE International Conference on Computer Vision, ICCV (pp. 22-29).

\bibitem{lu2017knowing}
Lu, J., Xiong, C., Parikh, D., \& Socher, R. (2017, July). Knowing when to look: Adaptive attention via a visual sentinel for image captioning. In Proceedings of the IEEE Conference on Computer Vision and Pattern Recognition (CVPR) (Vol. 6, p. 2).








\bibitem{anderson2018bottom}
Anderson, P., He, X., Buehler, C., Teney, D., Johnson, M., Gould, S., \& Zhang, L. (2018). Bottom-up and top-down attention for image captioning and visual question answering. In CVPR (Vol. 3, No. 5, p. 6).

\bibitem{zhang2018more}
Zhang, M., Yang, Y., Zhang, H., Ji, Y., Shen, H. T., \& Chua, T. S. (2018). More is Better: Precise and Detailed Image Captioning using Online Positive Recall and Missing Concepts Mining. IEEE Transactions on Image Processing.

\bibitem{park2018towards}
Park, C. C., Kim, B., \& Kim, G. (2018). Towards Personalized Image Captioning via Multimodal Memory Networks. IEEE Transactions on Pattern Analysis and Machine Intelligence.

\bibitem{wang2018image}
Wang, Cheng, Haojin Yang, and Christoph Meinel. "Image Captioning with Deep Bidirectional LSTMs and Multi-Task Learning." ACM Transactions on Multimedia Computing, Communications, and Applications (TOMM) 14.2s (2018): 40.

\bibitem{rennie2017self}
Rennie, S. J., Marcheret, E., Mroueh, Y., Ross, J., \& Goel, V. (2017, July). Self-critical sequence training for image captioning. In CVPR (Vol. 1, No. 2, p. 3).

\bibitem{kiros2014multimodal}
Kiros, Ryan, Ruslan Salakhutdinov, and Rich Zemel. "Multimodal neural language models." International Conference on Machine Learning. 2014.

\bibitem{wu2017image}
Wu, Q., Shen, C., Wang, P., Dick, A., \& van den Hengel, A. (2017). Image captioning and visual question answering based on attributes and external knowledge. IEEE transactions on pattern analysis and machine intelligence.

\bibitem{vinyals2015show}
Vinyals, Oriol, et al. "Show and tell: A neural image caption generator." Proceedings of the IEEE conference on computer vision and pattern recognition. 2015.

\bibitem{Karpathy2014Deep}
Karpathy, Andrej, Armand Joulin, and Fei Fei F. Li. "Deep fragment embeddings for bidirectional image sentence mapping." Advances in neural information processing systems. 2014.

\bibitem{Xu2015Show}
Xu, Kelvin, et al. "Show, attend and tell: Neural image caption generation with visual attention." International Conference on Machine Learning. 2015.

\bibitem{Fang2015captions}
Fang, Hao, et al. "From captions to visual concepts and back." Proceedings of the IEEE conference on computer vision and pattern recognition. 2015.

\bibitem{Karpathy2015Deep}
Karpathy, Andrej, and Li Fei-Fei. "Deep visual-semantic alignments for generating image descriptions." Proceedings of the IEEE Conference on Computer Vision and Pattern Recognition. 2015.

\bibitem{Anne2016Deep}
Anne Hendricks, Lisa, et al. "Deep compositional captioning: Describing novel object categories without paired training data." Proceedings of the IEEE Conference on Computer Vision and Pattern Recognition. 2016.

\bibitem{Chen2015Mind}
Chen, Xinlei, and C. Lawrence Zitnick. "Mind's eye: A recurrent visual representation for image caption generation." Proceedings of the IEEE conference on computer vision and pattern recognition. 2015.

\bibitem{Devlin2015Language}
Devlin, Jacob, et al. "Language models for image captioning: The quirks and what works." arXiv preprint arXiv:1505.01809 (2015).

\bibitem{Donahue2015Long-term}
Donahue, Jeffrey, et al. "Long-term recurrent convolutional networks for visual recognition and description." Proceedings of the IEEE conference on computer vision and pattern recognition. 2015.

\bibitem{Gan2017Stylenet}
Gan, Chuang, et al. "Stylenet: Generating attractive visual captions with styles." CVPR, 2017.

\bibitem{Jin2015Aligning}
Jin, Junqi, et al. "Aligning where to see and what to tell: image caption with region-based attention and scene factorization." arXiv preprint arXiv:1506.06272 (2015).

\bibitem{Kiros2014Unifying}
Kiros, Ryan, Ruslan Salakhutdinov, and Richard S. Zemel. "Unifying visual-semantic embeddings with multimodal neural language models." arXiv preprint arXiv:1411.2539 (2014).

\bibitem{Kiros2014multiplicative}
Kiros, Ryan, Richard Zemel, and Ruslan R. Salakhutdinov. "A multiplicative model for learning distributed text-based attribute representations." Advances in neural information processing systems. 2014. 

\bibitem{Mao2014deep}
Mao, Junhua, et al. "Deep captioning with multimodal recurrent neural networks (m-rnn)." arXiv preprint arXiv:1412.6632 (2014).

\bibitem{Memisevic2007Unsupervised}
Memisevic, Roland, and Geoffrey Hinton. "Unsupervised learning of image transformations." Computer Vision and Pattern Recognition, 2007. CVPR'07. IEEE Conference on. IEEE, 2007.

\bibitem{Pu2016Variational}
Pu, Yunchen, et al. "Variational autoencoder for deep learning of images, labels and captions." Advances in Neural Information Processing Systems. 2016.

\bibitem{Socher2014Grounded}
Socher, Richard, et al. "Grounded compositional semantics for finding and describing images with sentences." Transactions of the Association for Computational Linguistics 2 (2014): 207-218.

\bibitem{Sutskever2011Generating}
Sutskever, Ilya, James Martens, and Geoffrey E. Hinton. "Generating text with recurrent neural networks." Proceedings of the 28th International Conference on Machine Learning (ICML-11). 2011.

\bibitem{Sutskever2014Sequence}
Sutskever, Ilya, Oriol Vinyals, and Quoc V. Le. "Sequence to sequence learning with neural networks." Advances in neural information processing systems. 2014.

\bibitem{LTran2015Learning}
LTran, Du, et al. "Learning spatiotemporal features with 3d convolutional networks." Proceedings of the IEEE international conference on computer vision. 2015.

\bibitem{Tran2016Rich}
Tran, Kenneth, et al. "Rich image captioning in the wild." Proceedings of the IEEE Conference on Computer Vision and Pattern Recognition Workshops. 2016.

\bibitem{wu2016value}
Wu, Qi, et al. "What value do explicit high level concepts have in vision to language problems?." Proceedings of the IEEE Conference on Computer Vision and Pattern Recognition. 2016.

\bibitem{Yang2016Review}
Yang, Zhilin, et al. "Review networks for caption generation." Advances in Neural Information Processing Systems. 2016.

\bibitem{You2016Image}
You, Quanzeng, et al. "Image captioning with semantic attention." Proceedings of the IEEE Conference on Computer Vision and Pattern Recognition. 2016.

\bibitem{Young2014image}
Young, Peter, et al. "From image descriptions to visual denotations: New similarity metrics for semantic inference over event descriptions." Transactions of the Association for Computational Linguistics 2 (2014): 67-78.

\bibitem{Farhadi2010}
Farhadi, Ali, et al. "Every picture tells a story: Generating sentences from images." European conference on computer vision. Springer, Berlin, Heidelberg, 2010.

\bibitem{Gan2016}
Gan, Zhe, et al. "Semantic compositional networks for visual captioning." arXiv preprint arXiv:1611.08002 (2016).

\bibitem{Girshick2014}
Girshick, Ross, et al. "Rich feature hierarchies for accurate object detection and semantic segmentation." Proceedings of the IEEE conference on computer vision and pattern recognition. 2014.

\bibitem{Hodosh2013}
Hodosh, Micah, Peter Young, and Julia Hockenmaier. "Framing image description as a ranking task: Data, models and evaluation metrics." Journal of Artificial Intelligence Research47 (2013): 853-899.

\bibitem{Jia2015}
Jia, Xu, et al. "Guiding the long-short term memory model for image caption generation." Proceedings of the IEEE International Conference on Computer Vision. 2015.

\bibitem{Krishna2017}
Krishna, Ranjay, et al. "Visual genome: Connecting language and vision using crowdsourced dense image annotations." International Journal of Computer Vision 123.1 (2017): 32-73.

\bibitem{Kulkarni2013}
Kulkarni, Girish, et al. "Babytalk: Understanding and generating simple image descriptions." IEEE Transactions on Pattern Analysis and Machine Intelligence 35.12 (2013): 2891-2903.

\bibitem{Li2011}
Li, Siming, et al. "Composing simple image descriptions using web-scale n-grams." Proceedings of the Fifteenth Conference on Computational Natural Language Learning. Association for Computational Linguistics, 2011.

\bibitem{Kuznetsova2014}
Kuznetsova, Polina, et al. "TREETALK: Composition and Compression of Trees for Image Descriptions." TACL 2.10 (2014): 351-362.

\bibitem{Mao2015}
Mao, Junhua, et al. "Learning like a child: Fast novel visual concept learning from sentence descriptions of images." Proceedings of the IEEE International Conference on Computer Vision. 2015.

\bibitem{Mathews2016}
Mathews, Alexander Patrick, Lexing Xie, and Xuming He. "SentiCap: Generating Image Descriptions with Sentiments." AAAI. 2016.

\bibitem{Mitchell2012}
Mitchell, Margaret, et al. "Midge: Generating image descriptions from computer vision detections." Proceedings of the 13th Conference of the European Chapter of the Association for Computational Linguistics. Association for Computational Linguistics, 2012.

\bibitem{Ordonez}
Ordonez, Vicente, Girish Kulkarni, and Tamara L. Berg. "Im2text: Describing images using 1 million captioned photographs." Advances in Neural Information Processing Systems. 2011.

\bibitem{Yang2011}
Yang, Yezhou, et al. "Corpus-guided sentence generation of natural images." Proceedings of the Conference on Empirical Methods in Natural Language Processing. Association for Computational Linguistics, 2011.

\end{thebibliography}
\end{document}